\documentclass{article}
\usepackage{spconf,amsmath,graphicx,amsfonts,hhline,multirow}

\usepackage{color}

\DeclareMathOperator*{\argmax}{arg\,max}

\title{Abstractive Headline Generation for Spoken Content by Attentive Recurrent Neural Networks with ASR Error Modeling}

\name{Lang-Chi Yu$^1$, Hung-yi Lee$^2$ and Lin-shan Lee$^{12}$ \thanks{Copyright 2016 IEEE. Published in the 2016 IEEE Workshop on Spoken Language Technology (SLT 2016), scheduled for 13-16 December 2016 in San Juan, Puerto Rico. Personal use of this material is permitted. However, permission to reprint/republish this material for advertising or promotional purposes or for creating new collective works for resale or redistribution to servers or lists, or to reuse any copyrighted component of this work in other works, must be obtained from the IEEE. Contact: Manager, Copyrights and Permissions / IEEE Service Center / 445 Hoes Lane / P.O. Box 1331 / Piscataway, NJ 08855-1331, USA. Telephone: + Intl. 908-562-3966.}}
\address{
    $^1$Graduate Institute of Communication Engineering, National Taiwan University\\
    $^2$Graduate Institute of Computer Science and Information Engineering, National Taiwan
    University\\
    {\small \tt \{r04942056, hungyilee\}@ntu.edu.tw, lslee@gate.sinica.edu.tw}
}

\begin{document}
\maketitle
\begin{abstract}
Headline generation for spoken content is important since spoken content is difficult to be shown on the screen and browsed by the user. It is a special type of abstractive summarization, for which the summaries are generated word by word from scratch without using any part of the original content. Many deep learning approaches for headline generation from text document have been proposed recently, all requiring huge quantities of training data, which is difficult for spoken document summarization. In this paper, we propose an ASR error modeling approach to learn the underlying structure of ASR error patterns and incorporate this model in an Attentive Recurrent Neural Network (ARNN) architecture. In this way, the model for abstractive headline generation for spoken content can be learned from abundant text data and the ASR data for some recognizers. Experiments showed very encouraging results and verified that the proposed ASR error model works well even when the input spoken content is recognized by a recognizer very different from the one the model learned from.

\end{abstract}

\begin{keywords}
abstractive summarization, headline generation, ASR error modeling, 
attention mechanism, encoder-decoder architecture
\end{keywords}

\section{Introduction}
\label{sec:intro}

Document summarization is to generate a concise version of a given document while preserving the core information. This is important for both written and spoken content, because both of them usually include redundant, noisy, or less informative parts causing interference to users who wish to grasp the key information quickly. It is much more crucial for spoken content than for written content since spoken content is difficult to be shown on the screen and browsed by the user, while summaries of spoken content are very helpful in browsing. There are two categories for the summarization task. In \textit{extractive approaches}, the important parts of the original data are extracted and put together to form the summary. In contrast, in \textit{abstractive approaches}, the summary is generated word by word from scratch without using any part of the original content. When the abstractive summarization result includes only one sentence, this is usually referred to as \textit{sentence summarization}. Headline generation~\cite{banko2000headline,dorr2003hedge,xu2010keyword} is an example of abstractive sentence summarization, and is extremely important for spoken content, because with the headlines the users do not have to go through the lengthy part of the spoken content which they are not interested. We focus on abstractive headline generation for spoken content in this paper.

Abstractive summarization for text content has been successful with Deep Neural Network (DNN) techniques, for example those using DNN models with attention mechanism~\cite{rush2015neural,bahdanau2014neural} and using RNN models with encoder-decoder architecture~\cite{lopyrev2015generating,filippova2015sentence,chopraabstractive,gu2016incorporating,cheng2016neural,gulcehre2016pointing} useful in neural machine translation~\cite{cho2014learning} and dialogue model~\cite{shang2015neural}. Improved training techniques were also developed. For example, scheduled sampling~\cite{bengio2015scheduled} was applied to abstractive summarization~\cite{lopyrev2015generating} to bridge the gap between training and inference stage due to the differences in the input tokens to the decoder. The models can also be learned to directly optimize some evaluation metrics~\cite{ranzato2015sequence,shen2016neural}.

However, all the above works focused on text content, while such neural network based approaches for spoken content summarization were rarely seen, probably due to the difficulties in acquiring enough quantities of spoken content including the reference summaries to train such models. For example, in the previous works for text summarization, the training datasets were English Gigaword corpus \cite{graff2003english} for English and LCSTS corpus \cite{hu2015lcsts} for Chinese, which included respectively 4 million and 2.4 million document-headline pairs. To collect speech corpora including the reference summaries in the quantities of this order of magnitude is probably difficult. It is certainly possible to directly apply the transcriptions of audio data to the summarization models trained on text corpora, but the ASR errors would inevitably degrade the summarization performance, since these models never learned how to generate the abstract summaries from content with ASR errors.

In this paper, we solve this problem by developing an ASR error model learned from the ASR data for some recognizer and incorporate this model with an Attentive RNN (ARNN) encoder-decoder architecture, in order to learn from written content to generate headlines from spoken content. This paper is organized as follows: in Section \ref{sec:models}, we define the task, introduce the previously proposed architectures, and present the model proposed in this paper. We then describe the experimental setup in Section \ref{sec:experiments}, and present the results in Section \ref{sec:results} and concluding remarks in Section \ref{sec:conclusion}.

\section{Models}
\label{sec:models}

\subsection{Task Definition}
\label{ssec:task_def}

Our summarization task is defined as below. Given an input sequence $\textbf{X} = [x_{1},\ldots,x_{M}]$, which is a sequence of $M$ tokens from a fixed known dictionary $\mathcal{V_{X}}$, the model is to find $\textbf{Y} = [y_{1},\ldots,y_{N}]$, which is another sequence of $N$ tokens from another fixed known dictionary $\mathcal{V_{Y}}$. Here $\textbf{X}$ is the input text or spoken documents expressed as a sequence, and $\textbf{Y}$ is the abstractive headline expressing the meaning of $\textbf{X}$ in the most concise way. For example, in our experiments below, $\mathcal{V_{Y}}$ is the set of all allowed Chinese characters, and $\mathcal{V_{X}}$ can be the same character set as $\mathcal{V_{Y}}$, or a set of Initials and Finals of Mandarin. Initial is the initial consonant of a Mandarin syllable, while Final is the vowel part including optional medials and nasal ending. This is because the input spoken content can be expressed as a sequence of phonetic symbols. Also, since very often recognition errors are caused by incorrectly recognized phonetic units, expressing the input as a sequence of phonetic units may be helpful in ASR error modeling as will be clear below. The task here can then be considered with a conditional probability $P(Y\vert\textbf{X})$ for all possible $Y$ such that the desired output $\textbf{Y} = \argmax_{Y}P(Y\vert\textbf{X})$. This probability $P(Y\vert\textbf{X})$ is usually parameterized by a set of neural parameters $\theta$ as $P(Y\vert\textbf{X};\theta)$, and $\textbf{Y}$ is usually obtained sequentially by predicting every token in $\textbf{Y}$ based on the previous token,
\begin{equation} \label{eq:pred_factorized}
P(Y\vert\textbf{X};\theta) = \prod_{i=1}^{N} P(y_{i}\vert y_{1},\ldots,y_{i-1};\textbf{X};\theta),
\end{equation}
which can be modeled with RNN encoder-decoder architectures~\cite{cho2014learning,sutskever2014sequence,vinyals2015pointer,vinyals2015neural} described in the following.

\subsection{RNN Encoder-Decoder Architecture}
\label{ssec:rnn_arch}

\begin{figure}[t!]
\begin{minipage}[t]{1.0\linewidth}
  \centering
  \centerline{\includegraphics[width=8.5cm]{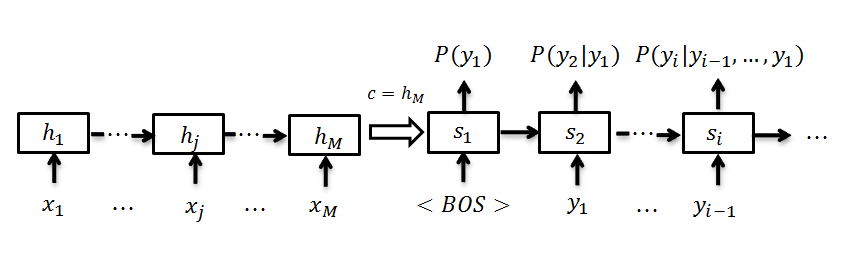}}
  \caption{RNN encoder-decoder architecture}
  \label{fig:rnn_arch}
\end{minipage}
\end{figure}

An RNN encoder-decoder architecture is shown in Fig.~\ref{fig:rnn_arch}. It consists of two parts: the encoder RNN and the decoder RNN. The encoder reads the input $\textbf{X}$ one token at a time and updates its hidden state $h_{j}$ according to the current input $x_{j}$ and the previous hidden state $h_{j-1}$, 
\[h_{j} = rnn_{e}(x_{j}, h_{j-1}), \quad j \in \{ 1, \ldots ,M \}\]
where $rnn_{e}(.,.)$ is a nonlinear function. After the encoder reads the last token $x_{M}$, it outputs a \textit{context vector} $c = h_{M} \in \mathbb{R}^{n}$ as the learned representation of the whole input sequence $\textbf{X}$.

The decoder then predicts $\textbf{Y}$ one token at a time given the context vector $c$ and all previous predicted tokens based on \eqref{eq:pred_factorized}. The conditional probability in \eqref{eq:pred_factorized} can be expressed as
\begin{equation} \label{eq:rnn_pred}
P(y_{i}\vert y_{1},\ldots,y_{i-1};\textbf{X};\theta) = dec(y_{i-1}, s_{i-1}, c=h_{M}),
\end{equation}
\begin{equation} \label{eq:rnn_h}
s_{i} = rnn_{d}(y_{i-1}, s_{i-1}, c=h_{M}),
\end{equation}
where $rnn_{d}(.,.,.)$ and $dec(.,.,.)$ are certain nonlinear functions, $i \in \{ 1, \ldots ,N \}$, $y_{0} =$ \textless BOS\textgreater, a special token for \textit{beginning of sentence}, and $s_{i} \in \mathbb{R}^{n}$ is the decoder RNN hidden state at $i$-th output step. The i-th output token $y_{i}$ is then the one which maximizes the probability in \eqref{eq:rnn_pred}, which is then used in \eqref{eq:rnn_pred}\eqref{eq:rnn_h} for decoding the next output token $y_{i+1}$. During training, the previous output token $y_{i-1}$ can be the labeled reference output. 

\subsection{Attentive RNN (ARNN) Architecture}
\label{ssec:att_arch}

\begin{figure}[t!]
\begin{minipage}[t]{1.0\linewidth}
  \centering
  \centerline{\includegraphics[width=8.5cm]{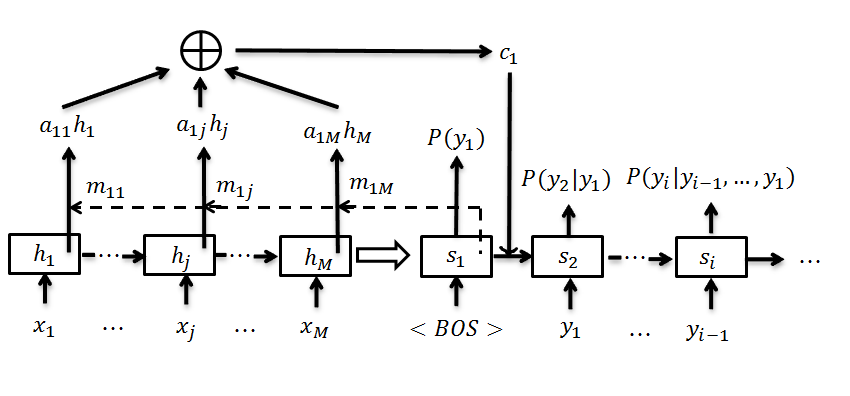}}
  \caption{Attentive RNN (ARNN) encoder-decoder architecture}
  \label{fig:att_arch}
\end{minipage}
\end{figure}

In the above architecture, the next token $y_{i}$ of $\textbf{Y}$ is predicted based on the conditional probability in \eqref{eq:rnn_pred}, which is determined by the context vector $c$ containing information of all the tokens in $\textbf{X}$. Nonetheless, not all tokens in $\textbf{X}$ are equally informative for the decoding process, and some of the input token may be noisy. An improved Attentive RNN architecture was then proposed as in Fig.~\ref{fig:att_arch}~\cite{bahdanau2014neural}. In Fig.~\ref{fig:att_arch}, the context vector $c$ in~\eqref{eq:rnn_pred} and~\eqref{eq:rnn_h} is modified as the weighted sum of the encoder hidden states at all input steps,
\begin{equation} \label{eq:att_weighted_sum}
c_{i} = \sum_{j=1}^{M} a_{ij}h_{j}.
\end{equation}
The weight $a_{ij}$ is defined as
\[a_{ij} = \frac{exp(m_{ij})}{\sum_{k=1}^{M} exp(m_{ik})},\]
where $m_{ij}$ is the cosine similarity between the decoder hidden state $s_{i}$ and encoder hidden state $h_{j}$. This implies that the input tokens better matched to the output token being decoded are given higher weights.

With the new context vector $c_{i}$ in~\eqref{eq:att_weighted_sum}, the decoding process in~\eqref{eq:rnn_pred}~\eqref{eq:rnn_h} are modified accordingly, 
\begin{equation} \label{eq:att_pred}
P(y_{i}\vert y_{1},\ldots,y_{i-1};\textbf{X};\theta) = dec(y_{i-1}, s_{i-1}, c_{i-1}),
\end{equation}
\begin{equation} \label{eq:att_h}
s_{i} = rnn_{d}(y_{i-1}, s_{i-1}, c_{i-1}).
\end{equation}
In this way, different input tokens are weighted differently for different output tokens, i.e., the decoder pays more \textit{attention} to those input tokens more useful for the output token it is currently decoding. In Fig.~\ref{fig:att_arch}, we only show the decoding process of $y_2$.

\subsection{ASR error confusion function}
\label{ssec:sim_asr}

The ASR error modeling can be started with a simplified confusion function. ASR errors can be considered as a transformation called \textit{confusion} here, $F: \textbf{X} \mapsto \textbf{X}'$, where $\textbf{X} = [x_{1},\ldots,x_{M}]$ is the correct input sequence and $\textbf{X}' = [x_{1}',\ldots,x_{M'}']$ the ASR results, both of which are sequences of tokens from the dictionary $\mathcal{V_{X}}$ in~\ref{ssec:rnn_arch}. We may approximate $F$ with a simplified context-independent confusion matrix trained with the output samples from a speech recognizer we wish to model. We first align the pairs of correct and ASR transcriptions with minimum Levenshtein distance using dynamic programming. With the alignment, we compute the confusion probability as
\begin{equation} \label{eq:confusion_prob}
P(q \vert p) = \frac{count(x' = q; x = p)}{\sum_{k} count(x' = k; x = p)}
\end{equation}
where $p, q, k$ are distinct token items in $\mathcal{V_{X}}$, and $count(x' = q; x = p)$ is the number of token $q$ in ASR results aligned to token $p$ in correct transcriptions. The summation in the denominator is over all allowed distinct token items in $\mathcal{V_{X}}$.

Now we can define $F$ as
\begin{equation} \label{eq:confused_process}
F(\textbf{X}) = [f(x_{1}),\ldots,f(x_{M})]
\end{equation}
where $f(x)$ is a distribution $\{P(k \vert x)$, k is a distinct token item in $\mathcal{V_{X}}\}$ over all allowed distinct token item $k$ in $\mathcal{V_{X}}$. So, for a given sequence $\textbf{X}$, the \textit{confused} sequence, $\textbf{X}' = F(\textbf{X}) = [x_{1}',\ldots,x_{M}']$ is stochastic because each $x_{j}'$ in $\textbf{X}'$ can be any token in $\mathcal{V_{X}}$ with some probability.

\subsection{ASR Error Modeling for ARNN}
\label{ssec:approaches}

There can be at least two possible approaches for error modeling with the Attentive RNN (ARNN) as explained below.

\subsubsection{Na\"ive Approach}
\label{ssec:naive}
We can simply apply $F(.)$ in~\eqref{eq:confused_process} to the input data $\textbf{X}$ to obtain many confused data $F(\textbf{X})$, and use these confused data and their headlines to train the models in Section~\ref{ssec:rnn_arch} and~\ref{ssec:att_arch}. We call this method \textit{na\"ive approach}.

\subsubsection{Proposed Approach} 
\label{ssec:proposed_approach}
In this approach, we modify the attention mechanism in Section~\ref{ssec:att_arch} with the function $F(\textbf{X})$ in~\eqref{eq:confused_process} in order to explore the underlying structure of ASR error patterns. First, we define a function
\begin{equation} \label{eq:f_likelihood}
e(\textbf{X}) = [P(x_{1}=f(x_{1})),\ldots,P(x_{M}=f(x_{M})] = [e_{1},\ldots,e_{M}]
\end{equation}
which is a vector of dimensionality $M$, the lengths of the input sequence $\textbf{X} = [x_{1},\ldots,x_{M}]$. The j-th element $e_{j}$ in~\eqref{eq:confusion_prob} is the probability that $x_{j}$ is correct. So this vector gives the likelihood whether or not a token in $\textbf{X}$ is unaffected by ASR. We apply \eqref{eq:f_likelihood} to \eqref{eq:att_weighted_sum}:
\begin{equation} \label{eq:att_weight_with_error}
c_{i} = \sum_{j=1}^{M} e_{j}a_{ij}h_{j}.
\end{equation}
In this way, the decoder pays more attention to those tokens that are more likely to be correct.

\begin{figure}[t!]
\begin{minipage}[t]{1.0\linewidth}
  \centering
  \centerline{\includegraphics[width=8.5cm]{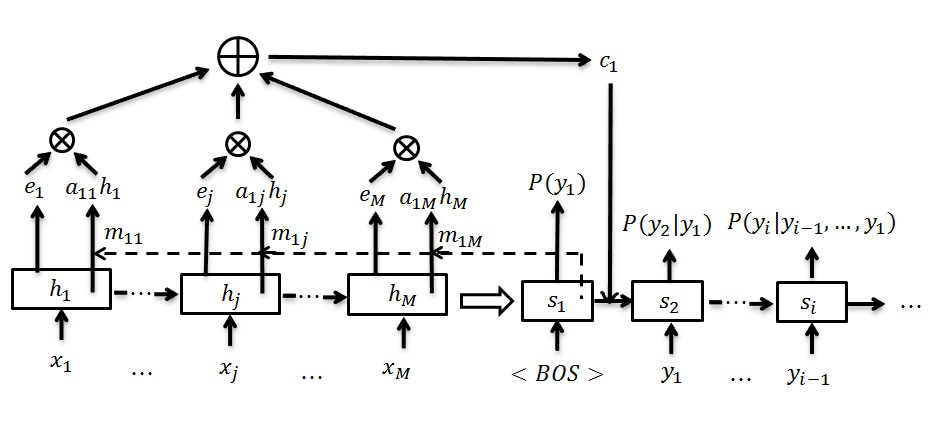}}
  \caption{The proposed ARNN with error modeling}
  \label{fig:proposed_model}
\end{minipage}
\end{figure}

To estimate the elements $e_{j}$ in $e(\textbf{X})$, we train a sequential error estimation model $see(.,.)$:
\begin{equation} \label{eq:error_estimation}
e_{j} = P(x_{j}=f(x_{j})) = see(x_{j}, h_{j-1}),
\end{equation}
where $see$ is a neural network whose training target can be easily obtained by comparing $\textbf{X}$ and any confused version of it. Note that an ASR error may lead to a significant change in semantics, which is why $e_{j}$ can be estimated sequentially as in~\eqref{eq:error_estimation}. This neural network and the attentive RNN can be jointly trained. In practice, $see$ is the direct output of the encoder RNN, which is unused in the architectures in Section~\ref{ssec:rnn_arch} and \ref{ssec:att_arch}, as shown in Fig.~\ref{fig:rnn_arch} and Fig.~\ref{fig:att_arch}. The complete attentive RNN with weighted attention in~\eqref{eq:att_weight_with_error} by error modeling is in Fig.~\ref{fig:proposed_model}. The training process includes the following steps:
\begin{enumerate}
\item We apply $F(.)$ in~\eqref{eq:confused_process} to each input training sequence $\textbf{X}$ to generate many different samples of $F(\textbf{X})$ since $F(\textbf{X})$ is stochastic.
\item The encoder reads all these confused sequences $F(\textbf{X})$ into hidden states $h_{1},\ldots,h_{M}$ and predicts the correctness $e_{j}$ for each input token $x_{j}$.
\item The decoder predicts $\textbf{Y}$ one token at a time based on the encoder hidden states and the weighted attention considering $e(\textbf{X})$ as in~\eqref{eq:att_weight_with_error}.
\end{enumerate}

In the testing process, step 1 above is skipped since the input is the ASR data. So Fig.~\ref{fig:proposed_model} actually shows the testing process. For training process, the input $x_{j}$ should be replaced by $f(x_{j})$.

\section{Experimental Setup}
\label{sec:experiments}

Here, we describe the corpora used and some implementation details.

\subsection{Datasets}
\label{ssec:datasets}

The ARNN model was trained on the Chinese Gigaword corpus~\cite{graff2005chinese}. This corpus consists of several years of news articles from Central News Agency of Taiwan and Xinhua News Agency of China. The following preprocessing steps were performed on this corpus. All Chinese characters were first converted into the traditional version of characters if they were not. Next, we removed articles from the time-period of the MATBN corpus (this corpus was used to train the ASR recognizer), replaced characters that occurred less than five times in the whole corpus with a special character \textless UNK\textgreater, and replaced Arabic numerals with \#~\cite{rush2015neural,cho2015using}. Note that the basic processing unit for the work here was the character, so there was no need to segment the character sequences into word sequences. In order to be able to take Initial/Final sequences as the input, we also converted the articles from character sequences to Initials/Final sequences with a pronunciation dictionary, which contained a total of 159 right-context-dependent Initials and context-independent Finals of Mandarin. After the preprocessing, we paired the first sentence of each news story with its headline to form a story-headline pair and removed those pairs whose headlines contained over 10\% of~\textless UNK\textgreater~symbols. The whole corpus was used on the training set, which consisted of about 2.07 million story-headline pairs and about 8K distinct characters.

The dataset used to obtain the confusion matrix for ASR error modeling and evaluation of the headline generation was the MATBN Mandarin Chinese broadcast news corpus \cite{wang2005matbn}. It contained a total of 198 hours of broadcast news from the Public Television Service Foundation of Taiwan with corresponding transcriptions, including human-generated headlines. We partitioned the corpus into two parts: 29K utterances for confusion matrix construction, and the rest 11k utterances for headline generation evaluation. For the part for evaluation, we paired the ASR results of each story with its corresponding headlines to form a story-headline pair. There are about 200 audio stories for the evaluation.

We used two different recognizers in the experiments here: the Kaldi toolkit \cite{povey2011kaldi}, and the online ASR recognizer Wit.ai \cite{witai}. For the recognizer with Kaldi toolkit, we used a tri-gram language model trained on 39M words of Yahoo news, and a set of acoustic models with 48 Gaussian mixtures per state and 3 states per model trained on a training corpus of 24.5 hours of Mandarin broadcast news different from MATBN. The character error rates (CER) for the MATBN corpus with Kaldi and Wit.ai were 28.72\% and 36.45\%, respectively. The confusion matrix used for ASR error modeling was obtained from the Kaldi toolkit, while the evaluation part was transcribed by both the Kaldi toolkit and Wit.ai. With the error modeling based on Kaldi toolkit performed on Wit.ai transcriptions, We wish to evaluate the robustness of the error modeling approach with respect to mismatched recognizers.

\subsection{Implementation}
\label{ssec:implementation}

We implemented the models with LSTM~\cite{hochreiter1997long} networks optimized by minimizing the negative log-likelihood between the predicted and the human-generated headlines with mini-batch stochastic gradient descent. The training setting summarized below were adjusted based on the validation set. 

The encoder and the decoder both had two hidden layers of 600 dimensions. The LSTM network parameters were initialized from a uniform distribution between $[-0.1, 0.1]$. The initial learning rate was 0.1, and divided by 1.15 if the log-likelihood in validation set did not improve for every 0.1 epoch. The training dropout rate was 5\%. Gradient clipping~\cite{pascanu2013difficulty} was adopted, with a gradient norm threshold of 10. The models were trained at most 7 epochs. During the training process, we adopted the scheduled sampling mechanism~\cite{bengio2015scheduled} with decay schedule of inverse sigmoid decay for $k=1$.

\section{Experimental Results}
\label{sec:results}

We evaluated the results with ROUGE-1, ROUGE-2, and ROUGE-L scores~\cite{lin2004rouge}. As mentioned in Section~\ref{ssec:task_def}, the input spoken content can be either character sequences or Initial/Final sequences. The models in Sections~\ref{ssec:rnn_arch} and \ref{ssec:att_arch} (RNN and ARNN) without ASR error modeling were taken as the baseline models. The \textit{na\"ive approach} of directly training the baseline models with confused input sequences described in Subsection~\ref{ssec:naive} and the proposed approach described in Subsection~\ref{ssec:proposed_approach} were compared.

\subsection{Oracle Results: Manual Transcriptions Input}
\label{ssec:baseline}

First, we tested the baseline models (RNN and ARNN) on manual transcriptions of news stories without ASR errors, which can be considered as the upper bound of the task. Table~\ref{tab:baseline} shows the results. The upper half of the table are for character sequence input (\textit{Char}), while the lower half for Initial/Final sequence input (\textit{I/F}). From Table~\ref{tab:baseline}, we observed the slight improvement obtained by including the attention mechanism (ARNN vs. RNN). Also, character sequence input performed significantly better than the Initial/Final sequence input in all cases. This is natural because in Chinese language there exist large number of homonym characters sharing the same pronunciation. So the pronunciation sequences carry much less information than character sequences.

      \begin{table}[th]
        \caption{\label{tab:baseline} 
        	{\it Oracle results: Baseline models for manual transcriptions input.}
        }
        \vspace{2mm}
        \centerline{
          \begin{tabular}{  l r | c | c | c | }
            \cline{3-5}
              & & ROUGE-1 & ROUGE-2 & ROUGE-L \\ \hline
              \multicolumn{1}{ |c  }{\multirow{2}{*}{Char} } &
              \multicolumn{1}{ |c| }{RNN} & 26.60 & 5.68 & 23.70 \\ \cline{2-5}
              \multicolumn{1}{ |c  }{} &
              \multicolumn{1}{ |c| }{ARNN} & 26.75 & 6.54 & 23.91 \\ \hhline{|=|=|=|=|=|}
              \multicolumn{1}{ |c  }{\multirow{2}{*}{I/F} } &
              \multicolumn{1}{ |c| }{RNN} & 21.78 & 3.72 & 19.75 \\ \cline{2-5}
              \multicolumn{1}{ |c  }{} &
              \multicolumn{1}{ |c| }{ARNN} & 22.22 & 4.00 & 20.17 \\ \hline
          \end{tabular}
        }
      \end{table}

\subsection{ASR Transcriptions Input}
\label{ssec:speech_summ}

The results for ASR transcriptions input obtained with Kaldi and Wit.ai are respectively in Table~\ref{tab:speech_summ_kaldi} and~\ref{tab:speech_summ_witai}. The upper half of Table~\ref{tab:speech_summ_kaldi} is for character sequence input. The baseline models (\textit{BSL}) in the rows (a)(b) refer to the same models as in Table~\ref{tab:baseline} (but in Table~\ref{tab:baseline}, ASR errors are not considered); the \textit{na\"ive} models (\textit{na\"i}) in rows (c)(d) refer to the na\"ive approaches proposed in  Section~\ref{ssec:naive}, i.e., baseline models but directly trained with confused data; and the proposed approach in row (e) is ARNN with error modeling. The lower half is the same but with Initial/Final sequence input. Table~\ref{tab:speech_summ_witai} is exactly the same, but with recognizer Wit.ai.

      \begin{table}[th]
        \caption{\label{tab:speech_summ_kaldi} 
        	{\it Results for ASR transcriptions input obtained with Kaldi.}
        }
        \vspace{2mm}
        \centerline{
        \resizebox{\columnwidth}{!}{
          \begin{tabular}{  l r r | c | c | c | }
            \cline{4-6}
              & & & ROUGE-1 & ROUEG-2 & ROUGE-L \\ \hline
              \multicolumn{1}{ |c  }{\multirow{5}{*}{Char} } &
                \multicolumn{1}{ |c  }{\multirow{2}{*}{BSL} } &
                  \multicolumn{1}{ |l| }{(a) RNN} & 21.87 & 4.93 & 20.52  \\ \cline{3-6}
                \multicolumn{1}{ |c|  }{} & \multicolumn{1}{ |c|  }{} &
                  \multicolumn{1}{ |l| }{(b) ARNN} & 21.32 & 4.84 & 20.05 \\ \cline{2-6}
                \multicolumn{1}{ |c  }{} &
                \multicolumn{1}{ |c  }{\multirow{2}{*}{na\"i} } &
                  \multicolumn{1}{ |l| }{(c) RNN} & 19.50 & 3.57 & 18.50  \\ \cline{3-6}
                \multicolumn{1}{ |c|  }{} & \multicolumn{1}{ |c|  }{} &
                  \multicolumn{1}{ |l| }{(d) ARNN} & 20.86 & 3.40 & 19.09 \\ \cline{2-6}
                \multicolumn{1}{ |c  }{} &
                \multicolumn{2}{ |c|  }{(e) Proposed} & \textbf{22.89} & \textbf{5.01} & \textbf{20.86} \\ \hhline{|=|==|=|=|=|}
                \multicolumn{1}{ |c  }{\multirow{5}{*}{I/F} } &
                \multicolumn{1}{ |c  }{\multirow{2}{*}{BSL} } &
                  \multicolumn{1}{ |l| }{(f) RNN} & 18.64 & 3.23 & 16.82  \\ \cline{3-6}
                \multicolumn{1}{ |c|  }{} & \multicolumn{1}{ |c|  }{} &
                  \multicolumn{1}{ |l| }{(g) ARNN} & 19.08 & 3.38 & 17.80 \\ \cline{2-6}
                \multicolumn{1}{ |c  }{} &
                \multicolumn{1}{ |c  }{\multirow{2}{*}{na\"i} } &
                  \multicolumn{1}{ |l| }{(h) RNN} & 16.87 & 2.36 & 15.42  \\ \cline{3-6}
                \multicolumn{1}{ |c|  }{} & \multicolumn{1}{ |c|  }{} &
                  \multicolumn{1}{ |l| }{(i) ARNN} & 17.14 & 2.42 & 16.06 \\ \cline{2-6}
                \multicolumn{1}{ |c  }{} &
                \multicolumn{2}{ |c|  }{(j) Proposed} & 20.67 & 3.66 & 18.83 \\ \hline
          \end{tabular}
        }
        }
      \end{table}

From rows (a)(b) of Table~\ref{tab:speech_summ_kaldi}, we see the baseline ARNN was actually slightly worse than baseline RNN (rows (b) vs. (a)), probably due to the wrong attention caused by ASR errors. In other words, the model paid attention to some tokens which were actually recognition errors. This situation is reversed in na\"ive approach (rows (d) vs. (c)), probably because the model may have learned to avoid to pay attention to incorrectly recognized errors. But the overall performance of na\"ive approach was worse than baseline (rows (c)(d) vs. (a)(b)), probably because the baseline models (rows (a)(b)) were trained with correct manual transcriptions, while the na\"ive models (rows (c)(d)) were trained with confused transcriptions and were therefore weaker. Having the error modeling telling the model which input tokens were more likely to be correct in the proposed approach (row (e)) as explained in Section~\ref{ssec:proposed_approach}, not only the wrong attention could be avoided, but the model learned how to take care of the errors when generating the headlines to a certain degree. So the performance of the model was much better (rows (e) vs. (a)(b)(c)(d)).

The lower half of Table~\ref{tab:speech_summ_kaldi} for Initial/Final sequence input offered lower performance just as in Table~\ref{tab:baseline}, but with similar trend as discussed above. The only difference was that here baseline ARNN was slightly better than baseline RNN (rows (g) vs. (f)). Because many homonym characters share the same pronunciation, the character error rate was much higher than the Initial/Final error rate. So the lower Initial/Final error rate led to much less wrong attention on recognition errors.

      \begin{table}[th]
        \caption{\label{tab:speech_summ_witai} 
        	{\it Results for ASR transcriptions input obtained with Wit.ai.}
        }
        \vspace{2mm}
        \centerline{
        \resizebox{\columnwidth}{!}{
          \begin{tabular}{  l r r | c | c | c | }
            \cline{4-6}
              & & & ROUGE-1 & ROUGE-2 & ROUGE-L \\ \hline
              \multicolumn{1}{ |c  }{\multirow{5}{*}{Char} } &
                \multicolumn{1}{ |c  }{\multirow{2}{*}{BSL} } &
                  \multicolumn{1}{ |l| }{(a) RNN} & 20.07 & 2.98 & 18.44  \\ \cline{3-6}
                \multicolumn{1}{ |c|  }{} & \multicolumn{1}{ |c|  }{} &
                  \multicolumn{1}{ |l| }{(b) ARNN} & 16.33 & 2.14 & 15.23 \\ \cline{2-6}
                \multicolumn{1}{ |c  }{} &
                \multicolumn{1}{ |c  }{\multirow{2}{*}{na\"i} } &
                  \multicolumn{1}{ |l| }{(c) RNN} & 19.58 & 2.91 & 18.54  \\ \cline{3-6}
                \multicolumn{1}{ |c|  }{} & \multicolumn{1}{ |c|  }{} &
                  \multicolumn{1}{ |l| }{(d) ARNN} & 19.87 & 3.22 & 18.47\\ \cline{2-6}
                \multicolumn{1}{ |c  }{} &
                \multicolumn{2}{ |c|  }{(e) Proposed} & \textbf{20.40} & \textbf{3.44} & \textbf{18.65} \\ \hhline{|=|==|=|=|=|}
                \multicolumn{1}{ |c  }{\multirow{5}{*}{I/F} } &
                \multicolumn{1}{ |c  }{\multirow{2}{*}{BSL} } &
                  \multicolumn{1}{ |l| }{(f) RNN} & 10.14 & 0.57 & 9.98  \\ \cline{3-6}
                \multicolumn{1}{ |c|  }{} & \multicolumn{1}{ |c|  }{} &
                  \multicolumn{1}{ |l| }{(g) ARNN} & 10.82 & 0.45 & 10.72 \\ \cline{2-6}
                \multicolumn{1}{ |c  }{} &
                \multicolumn{1}{ |c  }{\multirow{2}{*}{na\"i} } &
                  \multicolumn{1}{ |l| }{(h) RNN} & 11.18 & 1.09 & 10.91  \\ \cline{3-6}
                \multicolumn{1}{ |c|  }{} & \multicolumn{1}{ |c|  }{} &
                  \multicolumn{1}{ |l| }{(i) ARNN} & 9.84 & 0.12 & 9.78 \\ \cline{2-6}
                \multicolumn{1}{ |c  }{} &
                \multicolumn{2}{ |c|  }{(j) Proposed} & 11.88 & 0.86 & 11.72 \\ \hline
          \end{tabular}
        }
        }
      \end{table}

The results using Wit.ai as the recognizer are listed in Table~\ref{tab:speech_summ_witai}, in which the confusion matrix used for error modeling was obtained by the transcriptions of Kaldi toolkit. Compared with the results in Table~\ref{tab:speech_summ_kaldi}, we see the scores in Table~\ref{tab:speech_summ_witai} are in general lower than those in Table~\ref{tab:speech_summ_kaldi}, not only because of the mismatched recognizers and recognition error patterns, but because the character error rate of Wit.ai was much higher than that of Kaldi (36.45\% vs. 28.72\%). The specially low performance of baseline ARNN (row (b)) was obviously because the very high character error rate caused too much wrong attention and disturbed the model. The very low performance of Initial/Final sequence input (lower half of Table~\ref{tab:speech_summ_witai}) indicated that the phonetic sequences with low accuracy carried too little information to be used for headline generation. However, we found that the proposed approach still performed very well (row (e)) for character sequence input even with the low ASR accuracy and the mismatched recognizers.

\section{Conclusion}
\label{sec:conclusion}

In this paper, we propose a novel attentive RNN (ARNN) architecture with ASR error modeling for headline generation for spoken content, which can be trained without a large corpus of speech-headline pairs. Experimental results show that the proposed model is able to learn the recognition error patterns and avoid the errors when paying attention to important tokens in generating the headlines. The model is even reasonably robust with respect to the mismatched condition that the input spoken content is recognized by a recognizer very different from the one the model learned from.

\bibliographystyle{IEEEbib}
\bibliography{refs}

\end{document}